\documentclass[letterpaper, 10pt, conference]{ieeeconf}

\IEEEoverridecommandlockouts 
\usepackage{graphics}
\usepackage{graphicx}
\usepackage{amsmath}
\usepackage{amssymb}
\usepackage{algorithm}
\usepackage{algorithmic}
\usepackage{booktabs}
\usepackage{multirow}
\usepackage{xcolor}
\usepackage{hyperref}
\usepackage{balance}
\usepackage{colortbl}
\usepackage{cite}
\DeclareMathOperator*{\argmax}{arg\,max}

\title{\LARGE \bf
ART-VS: Adaptive Resolution Tiling for Vision Transformer Visual Servoing
}

\author{Alessandro Scherl$^{1,2}$ \quad Bernhard Neuberger$^{2,3}$ \quad Simon Schwaiger$^{2,4}$\\David Mulero-Pérez$^{1}$ \quad Lucas Muster$^{2,5}$ \quad José García-Rodríguez$^{1}$
\thanks{$^{1}$Department of Computer Technology, University of Alicante, Spain. {\tt\small(email: as358@alu.ua.es})}%
\thanks{$^{2}$Department of Industrial Engineering, UAS Technikum Vienna, Austria.}%
\thanks{$^{3}$Automation and Control Institute, TU Wien, Vienna, Austria.}%
\thanks{$^{4}$Institute of Software Engineering and Artificial Intelligence, Graz University of Technology, Austria.}%
\thanks{$^{5}$Institute for Integrative Nature Conservation Research, University of Natural Resources and Life Sciences Vienna, Austria.}%
}
\begin{document}

\maketitle
\thispagestyle{empty}
\pagestyle{empty}

\begin{abstract}

Visual servoing with self-supervised Vision Transformer (ViT) features enables training-free robotic positioning with strong generalization, but faces a fundamental trade-off between robustness and precision. 
Coarse patch-level descriptors provide stable correspondences yet limit positioning accuracy.
Increasing image resolution improves precision but yields only marginal robustness gains - under perturbation, high-resolution processing improves convergence success rate from 76.6\% to just 81.0\% despite 12× more ViT patches.
Therefore, we propose Adaptive Resolution Tiling Visual Servoing (ART-VS), a two-phase method that adapts feature granularity to servoing progress: a coarse phase at native ViT resolution for stable alignment, then a tiled high-resolution phase that restricts matching to local neighborhoods improving positioning accuracy.
Without any task-specific training, ART-VS achieves 95.4\% convergence under perturbation, outperforming standard and full-resolution ViT-based servoing by 18.8 and 14.4 percentage points.
Over the former it reduces positioning error by 53\%, while running at over 10× higher speed and 27\% lower VRAM than the latter.
We validate ART-VS across three ViT backbones and demonstrate real-world category-level grasping of unseen object instances, achieving 95/100 on transparent bottles and 98/100 on shoes. Code available under \url{https://art-vs.github.io/}.

\end{abstract}

\section{INTRODUCTION}

Visual Servoing (VS) uses visual feedback to control robot motion toward a desired pose defined by a reference image~\cite{Chaumette2006, Chaumette2007}, and is widely applied in end-effector positioning and object manipulation tasks such as grasping~\cite{Puang2020, LaAnh2012}.
As real-world applications require generalization across diverse objects and scenes, central challenges arise: classical feature-based methods achieve high precision but are sensitive to occlusions and illumination changes~\cite{Karami2017}.
Deep learning approaches improve robustness at the cost of task-specific training, hence sacrificing general applicability~\cite{Felton2022, ICRA2023Felton, ICRA2018Bateux}.
Self-Supervised-Learning (SSL) Vision Transformer (ViT) features provide a solution: models such as DINOv2~\cite{DINOv2} produce patch-level descriptors that generalize across object instances without task-specific data.
Combined with classical Image Based Visual Servoing (IBVS) control, such features achieve high convergence rates under ideal conditions, but degrade under appearance perturbations~\cite{ViTVS2025}.
Recent ViT approaches support full input resolution~\cite{DINOv3}, yielding finer features that, as we demonstrate, improve positioning precision, yet two challenges remain:

\begin{figure}[t!]
    \centering
    \includegraphics[width=1\linewidth]{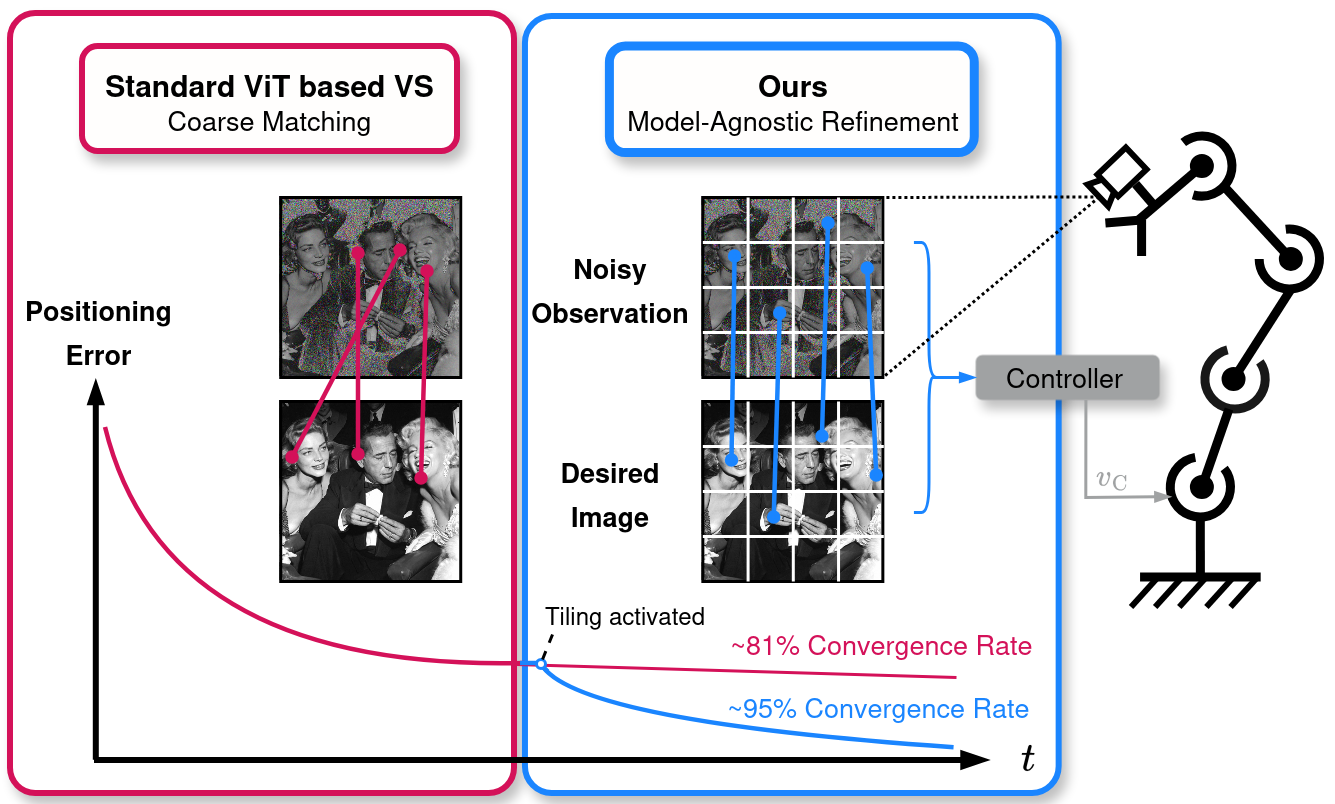}
    \caption{\textbf{ART-VS overview.} Left: During the coarse phase, standard ViT-based VS at native resolution provides stable correspondences despite large pose differences. Right: After sufficient error reduction, the tiled refinement phase increases effective resolution through spatially constrained matching, allowing for higher convergence rates.
    Bottom: Positioning error over-time.}
    \label{fig:teaser}
    \vspace{-2ex}
\end{figure}

\begin{enumerate}
    \item Robustness: 
    ViTs employ the full image for feature matching, thus allowing long-range matches, which is detrimental to convergence at finer ranges.
    \item Precision at scale: While finer features improve positioning accuracy, the quadratic self-attention cost of ViTs makes full-resolution processing prohibitively slow for closed-loop control.

\end{enumerate}

To alleviate the aforementioned problems, we present Adaptive Resolution Tiling Visual Servoing (ART-VS), a two-phase method consisting of a coarse phase and a tiled refinement phase.
During the coarse phase, ART-VS operates at the resolution the ViT was trained on, where patch-level aggregation provides stable correspondences for initial convergence.
Once the average matching error drops below a threshold~$\tau$ of the initial error, ensuring reliable tile-local matching (Sec.~\ref{sec:ablations}), the system transitions to the tiled refinement phase.
As illustrated in Fig.~\ref{fig:teaser}, the input is divided into a grid of non-overlapping tiles where matches are only allowed within the corresponding ones.
Consequently, ViT is forced to match features in narrow vicinities, thus reducing long-range mismatches and increasing the convergence rate.
In addition, tiles are matched independently in parallel, enabling full-resolution processing with shorter processing times.
ART-VS achieves both robust convergence and precise positioning without any task-specific training or scene-dependent tuning.
To enable deployment in cluttered scenes, we further propose an optional language-guided Region of Interest (ROI) detection and tracking mode that isolates the target object before servoing, using only an open-vocabulary prompt. 
We evaluate ART-VS across three ViT backbones in simulation and demonstrate practical applicability through robotic end-effector positioning and category level grasping experiments.

Our contributions are:
\begin{enumerate}
    \item A training-free, backbone-agnostic method for VS based on adaptive image tiling.
    Coarse patch aggregation provides robustness to appearance variation, while tiled refinement restricts matching to local, non-redundant neighborhoods, jointly improving convergence behavior and scaling.
    Evaluation across three ViT backbones, with the best achieving 95.4\% convergence rate under perturbation, improving over standard ViT-based servoing (18.8\,pp) and full-resolution ViT processing (14.4\,pp) at over 10$\times$ higher effective frame rate and 27\% lower VRAM.
    \item A VS-system for language-guided category-level grasping from a single reference image, requiring no task-specific training. Validated on 200 real-world trials across unseen object instances in cluttered multi-object scenes, achieving 95\% and 98\% success on transparent bottles and shoes, respectively.
\end{enumerate}

\section{RELATED WORK}

We review prior work along three axes: classical and learning-based servoing (Sec.~\ref{sec:rw_vs}), self-supervised ViT features for training-free generalization (Sec.~\ref{sec:rw_vit}), and resolution enhancement strategies that address their remaining spatial limitations (Sec.~\ref{sec:rw_resolution}).

\subsection{Visual Servoing}
\label{sec:rw_vs}

VS is broadly categorized into Position-Based Visual Servoing (PBVS), operating on estimated pose differences, and Image-Based (IBVS), minimizing error directly in image space~\cite{Chaumette2006, Chaumette2007, Hutchinson1996}.
Classical IBVS uses hand-crafted descriptors such as SIFT~\cite{SIFT}, ORB~\cite{ORB}, and AKAZE~\cite{AKAZE} for correspondence-based control~\cite{Hoffmann2006, LaAnh2012}, achieving sub-millimeter precision under controlled conditions but degrading under occlusions and illumination changes~\cite{Karami2017}.
Learning-based approaches improve robustness through CNN-based pose regression~\cite{ICRA2018Bateux}, Siamese networks~\cite{Felton2021}, unsupervised autoencoders~\cite{Felton2022}, and sim-to-real transfer~\cite{Puang2020} - yet encode scene-specific appearance, limiting generalization.
A complementary line of work learns neural control policies from keypoint correspondence graphs~\cite{CNS2024, DepthPC2025}, replacing the classical control law while relying on instance-level handcrafted detectors (e.g., SIFT) for feature extraction.

\subsection{Vision Transformer Features for Dense Correspondence}
\label{sec:rw_vit}

SSL Vision Transformers~\cite{Dosovitskiy2021} produce patch-level features with strong semantic consistency across instances.
Successive models such as DINO~\cite{Caron2021}, DINOv2~\cite{DINOv2}, and DINOv3~\cite{DINOv3} progressively improved zero-shot transfer and feature quality, while AM-RADIO~\cite{AMradio} distills multiple foundation models into a single backbone.
Amir~et~al.~\cite{amir2022deep} showed that spatially binned ViT descriptors serve as effective dense visual descriptors, and Goodwin~et~al.~\cite{Goodwin2022} applied cyclic consistency for zero-shot pose estimation.
In visual servoing, ViT-VS~\cite{ViTVS2025} combined DINOv2 features with cyclic matching and IBVS control, achieving high convergence rates without task-specific training, though with reduced robustness under appearance perturbations and positioning precision bounded by both the patch grid and the backbone's fixed input resolution.

\subsection{Spatial Refinement for Dense Correspondence}
\label{sec:rw_resolution}

To circumvent the problem of low feature density, learned feature up-sampling methods~\cite{fu2024featup, huang2025loftup} increase the spatial resolution of ViT features post-hoc through self-supervised objectives, but can only interpolate sub-patch detail already aggregated by the encoder's patch embedding.
Alternatively, coarse-to-fine matching is a well-established paradigm in correspondence estimation~\cite{melekhov2019dgc, sun2021loftr, sarlin2020superglue}. 
However, spatial feature refinement methods are typically trained with task-specific correspondence objectives within a given target domain.
While they may generalize outside of their training distribution, they represent a different trade-off in data dependence and deployment flexibility compared to purely foundation-model-based methods.
To retain the advantages of spatial feature refinement w.r.t. precision and stability in a training-free setting, we present adaptive resolution tiling.
Instead of forcing higher feature density through upscaled input images, as is possible with recent SSL ViT \cite{DINOv3}, we propose backbone-agnostic spatial and semantic matching in tiles.
This eases computational complexity while simultaneously surpressing global mismatches \cite{ViTVS2025} through spatial alignment, especially under perturbed query images.
Our method achieves this while preserving the zero-shot, foundation-model-based generality of ViT features by focusing on fully training-free pipelines.

\begin{figure*}[!t]
    \centering
    \includegraphics[width=0.92\textwidth]{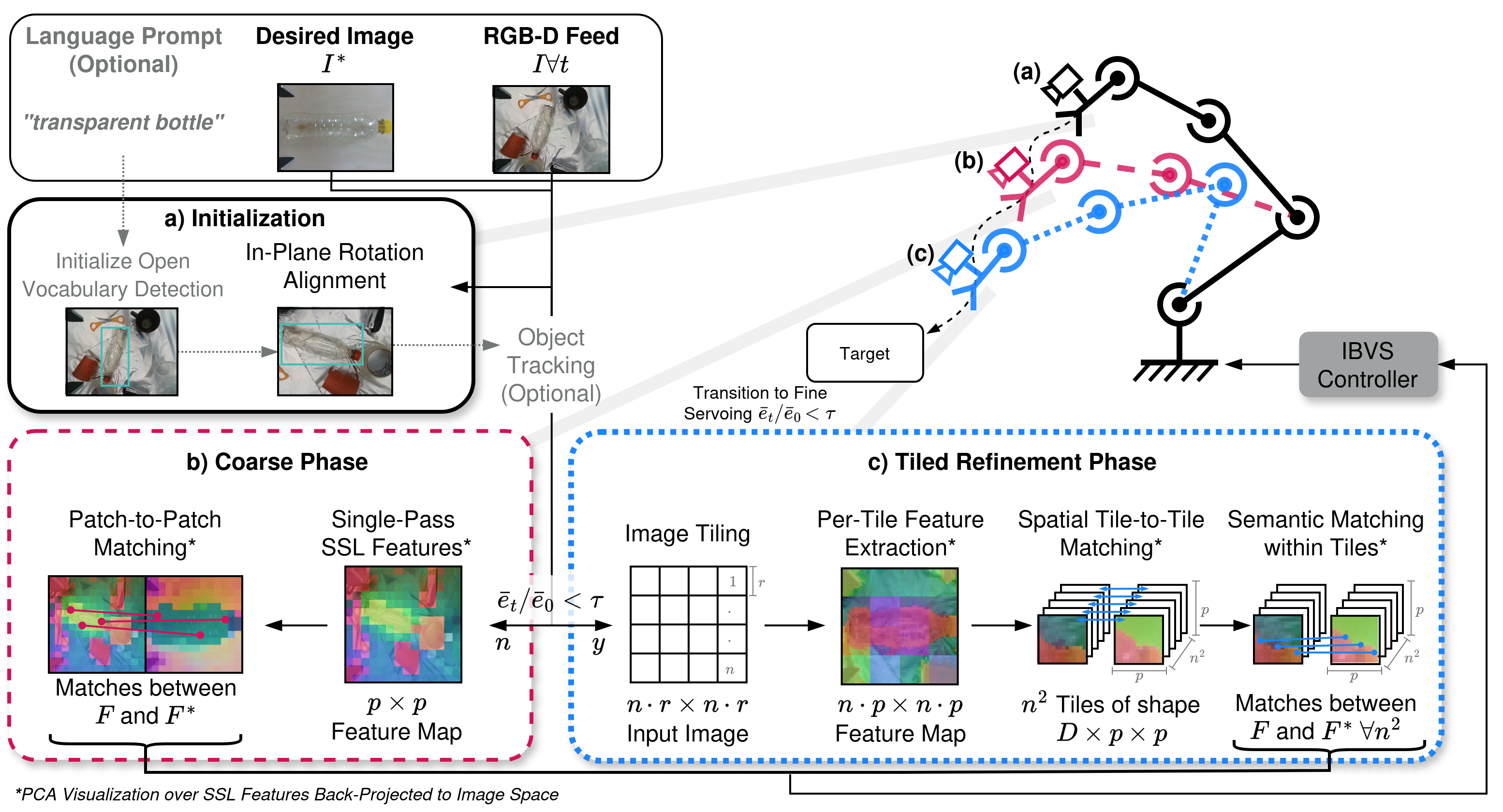}
    \caption{\textbf{ART-VS pipeline overview.} Inputs: an optional language prompt, desired image~$I^*$, and RGB-D feed~$I\forall t$. \textbf{a)~Initialization:} a language prompt triggers ROI-guided detection and tracking; otherwise full-image processing is used. In-plane rotation alignment is applied in both modes. The transition criterion~$(\bar{e}_t / \bar{e}_0 < \tau)$ selects between: \textbf{b)~Coarse Phase} - single-pass ViT encoding produces $p \times p$ feature maps with global patch-to-patch matching between~$\mathbf{F}$ and~$\mathbf{F}^*$; and \textbf{c)~Tiled Refinement Phase} - the input is partitioned into an adaptive $n \times n$ grid of $r \times r$ tiles matching the backbone's native resolution, each encoded independently, with matching restricted to spatially corresponding tiles between~$\mathbf{F}$ and~$\mathbf{F}^*$.}
    \label{fig:Methodology}
    \vspace{-2ex}
\end{figure*}

  \section{METHOD}
  \label{sec:method}

Fig.~\ref{fig:Methodology} provides an overview of ART-VS.
Given a desired RGB image~$I^*$ defining the desired pose and a current RGB-D stream~$I\forall t$, ART-VS performs VS in two phases that exploit complementary properties of ViT features at different spatial resolutions: coarse descriptors provide stable correspondences under large pose differences, while fine-grained features improve precision but are susceptible to long-range mismatches when the viewpoint gap is still large.
Both phases produce feature correspondences for a shared IBVS controller (Sec.~\ref{sec:preliminaries}).
After an initialization step that resolves in-plane rotation and optionally isolates the target object through language-guided ROI detection and tracking (Sect.~\ref{sec:initialization}), the system enters the coarse phase at native ViT resolution (Sect.~\ref{sec:coarse_phase}).
Once the error has been sufficiently reduced, the system transitions to the tiled high-resolution refinement phase with spatially constrained tile-local matching (Sec.~\ref{sec:tiling}).

  \subsection{Preliminaries}
  \label{sec:preliminaries}

  \textbf{Image-Based Visual Servoing.}
  We build upon the IBVS framework~\cite{Chaumette2006}.
  Given $K$ feature correspondences between the current and desired images, the feature error
  \begin{equation}
      \mathbf{e} = \mathbf{s} - \mathbf{s}^*
      \label{eq:error}
  \end{equation}
  measures the displacement between current feature locations $\mathbf{s} \in \mathbb{R}^{2K}$ and desired locations $\mathbf{s}^* \in \mathbb{R}^{2K}$, where each of the $K$ correspondences contributes a 2D pixel coordinate~$(u,v)$.
  The camera velocity
  \begin{equation}
      \mathbf{v}_c = -\lambda \, \mathbf{L}_s^{+} \, \mathbf{e}
      \label{eq:control_law}
  \end{equation}
  drives the error toward zero, where $\mathbf{L}_s^{+}$ is the pseudoinverse of the interaction matrix constructed from depth readings and $\lambda$ is the control gain.

  \textbf{ViT Feature Extraction.}
We employ a pretrained Vision Transformer as feature extractor~$f_v$, mapping an image of shape $\mathbb{R}^{3 \times r \times r}$ to a dense descriptor grid $\mathbf{F} \in \mathbb{R}^{D \times p \times p}$, where $r$ denotes the backbone's native input resolution, i.e.\ the spatial dimension for which the model's positional embeddings are defined (e.g.\ $r{=}256$ for DINOv3-S, $r{=}224$ for DINOv2-S), $p_s$ is the patch size, $p = r / p_s$ the resulting grid dimension, and $D$ the descriptor dimensionality.
The CLS token and, when present, register tokens are discarded; the embedding is retrieved from the remaining spatial patch tokens.
The current and desired images are encoded as $\mathbf{F} = f_v(I)$ and $\mathbf{F}^* = f_v(I^*)$, respectively.

\textbf{Best-Buddy Matching.}
  We establish correspondences via best-buddy matching~\cite{CVPR2015Dekel}.
  Given descriptor sets $\mathbf{F}^* = \{\mathbf{f}_1^*, \dots, \mathbf{f}_N^*\}$ and $\mathbf{F} = \{\mathbf{f}_1, \dots, \mathbf{f}_N\}$ with $N = p^2$, the set of mutually consistent correspondences is
  \begin{align}
      \mathcal{M} = \bigl\{(i,j) \;\big|\;
          & j = \argmax_k \cos(\mathbf{f}_i^*, \mathbf{f}_k) \notag\\
          \text{and}\;\;
          & i = \argmax_l \cos(\mathbf{f}_l^*, \mathbf{f}_j)
      \bigr\}
      \label{eq:best_buddy}
  \end{align}
where $\cos(\cdot,\cdot)$ denotes cosine similarity.
The first condition finds the most similar current patch~$j$ for each desired patch~$i$; the second requires that $j$'s closest match in the desired image is~$i$ itself, ensuring mutual consistency.
From~$\mathcal{M}$, $K$~correspondences are randomly subsampled per iteration to ensure spatial coverage across the descriptor grid~\cite{ViTVS2025}.

\subsection{Initialization}
\label{sec:initialization}

\textbf{In-Plane Rotation Alignment.}
As ViT patch descriptors are not rotation invariant, the in-plane offset between views is resolved by evaluating feature similarity at four discrete angles $\{0^\circ, 90^\circ, 180^\circ, 270^\circ\}$ and applying the best-matching rotation to the robot~\cite{ViTVS2025}.

\textbf{Language-Guided ROI (Optional).}
When a text prompt is provided (e.g., ``transparent bottle''), ART-VS restricts processing to the target object.
A language-promptable segmentation model generates a mask from the desired image, suppressing background patches during feature matching, while a single-object tracker propagates the bounding box across subsequent frames with re-detection on failure.
We use LangSAM~\cite{xu2025langsam} and LightTrack~\cite{LightTrack}, though the pipeline is agnostic to the specific components (Sec.~\ref{sec:ablations}).
When active, both servoing phases operate within the tracked region, and the tiling grid adapts to the bounding box aspect ratio, maximizing feature density on the target.

\subsection{Coarse Phase}
\label{sec:coarse_phase}

During the coarse phase, the current and desired images are each processed by~$f_v$ at the backbone's input resolution~$r$ in a single forward pass, and correspondences are established via global best-buddy matching~\eqref{eq:best_buddy} over the full
$p \times p$ descriptor grid.
The resulting $K$ correspondences drive the IBVS controller~\eqref{eq:control_law}.
We monitor the mean feature displacement
  \begin{equation}
      \bar{e} = \frac{1}{K} \sum_{i=1}^{K}
          \lVert \mathbf{s}_i - \mathbf{s}_i^* \rVert
      \label{eq:mean_error}
  \end{equation}
where $\mathbf{s}_i, \mathbf{s}_i^* \in \mathbb{R}^2$ are the pixel locations of the $i$-th correspondence and $\bar{e}_0$ is the value measured at the first servoing iteration after rotation alignment.
The coarse phase terminates when $\bar{e}_t / \bar{e}_0 < \tau$, with $\tau = 0.20$ (i.e., 80\% error reduction).
Beyond this point, the remaining pose difference is small enough for reliable tile-local matching, while the coarse descriptor grid limits further refinement, motivating the transition to higher-resolution processing (Sec.~\ref{sec:ablations}).

\subsection{Tiled Refinement Phase}
\label{sec:tiling}

After the coarse phase, ART-VS transitions to the tiled refinement phase, which increases feature resolution while constraining matches to local neighborhoods.

\textbf{Tiling and Feature Extraction.}
To maximize feature density for a given backbone, both images are partitioned into non-overlapping tiles, each processed at the backbone's input resolution~$r$.                                                                                          
For full-image tiling, the grid dimension is            
\begin{equation}                                       
     n = \max\!\left(                                    
        \left\lceil \frac{W}{r} \right\rceil,\;
        \left\lceil \frac{H}{r} \right\rceil
    \right)
    \label{eq:grid}
\end{equation}
where $W \times H$ is the original image resolution (e.g., $n{=}6$ for $r{=}256$ with $1440{\times}1080$ images).
Both images are resized to $(n \cdot r) \times (n \cdot r)$ and partitioned into $n^2$ non-overlapping tiles of $r \times r$ pixels, preserving full spatial detail per tile.
Non-overlapping partitioning ensures each pixel belongs to exactly one tile, preventing duplicate correspondences.
In ROI-guided mode (Sec.~\ref{sec:initialization}), the grid instead adapts to the bounding box aspect ratio with independent dimensions $n_w = \lceil W_{\mathrm{roi}} / r \rceil$ and $n_h = \lceil H_{\mathrm{roi}} / r \rceil$, yielding $n_w
\!\times\! n_h$ tiles (e.g., $2{\times}4$ for a tall object).
The resulting tiles from both images are processed by~$f_v$ in a single batched forward pass, yielding descriptor tensors $\mathbf{F},\, \mathbf{F}^* \in \mathbb{R}^{n^2 \times D \times p \times p}$ (or $\mathbb{R}^{n_w n_h \times D \times p \times
  p}$ in ROI mode).
    Desired-image descriptors are cached after the first extraction.

\textbf{Spatially Constrained Matching.}
During tiled inference, the ViT only attends across patches within the same tile. While SSL objectives encourage context-invariant embeddings, independent encoding provides no guarantee that descriptors remain comparable across tiles with different spatial context.
We therefore constrain matching so that each desired tile is matched exclusively against the current tile at the same grid position via best-buddy matching~\eqref{eq:best_buddy}.
This assumes that the preceding coarse phase has reduced the pose difference sufficiently for corresponding content to be co-located within the same tile across views.
Each tile contributes at most $K_t = \max(4,\, \lfloor K / n^2 \rfloor)$ randomly subsampled mutual nearest neighbors, ensuring that the final selection draws from spatially distributed regions.
Valid correspondences from all $n^2$ tile pairs are pooled, and the top~$K$ by cosine similarity are selected for the IBVS controller.

\textbf{Coordinate Recovery.}
Tile-local patch coordinates are mapped back to the original image frame by offsetting each patch position by its tile's location in the grid and rescaling from the tiling resolution~$n \cdot r$ to the original dimensions~$W \times H$.

\section{EXPERIMENTS}
\label{sec:experiments}

\begin{table*}[!h]
\vspace{1.5ex}
\caption{\textbf{Simulation benchmark results}.
$^\dagger$Results from~\cite{ICRA2023Felton},
Among patch-level ViT methods: \textbf{bold} = best under perturbation, \underline{underline} = best unperturbed.
ART-VS FPS values are effective FPS (see Sec.~\ref{sec:results}).}
\label{tab:hollywood}
\centering
\begin{tabular}{l|c|c|c|c|c|c}
\hline
\textbf{Method} & \textbf{Perturbed} & \textbf{Converged [\%]} & \textbf{Position [cm]} & \textbf{Orientation [°]} & \textbf{VRAM [MB]} & \textbf{FPS} \\
\hline\hline
\multicolumn{7}{l}{\textit{Scene-specific finetuning}} \\
\hline
PBVS-CNN~\cite{ICRA2018Bateux}$^\dagger$ & $\times$ & 75.6 & $3.35\pm0.65$ & $1.71\pm0.65$ & -- & -- \\
PBVS-CNN~\cite{ICRA2018Bateux}$^\dagger$ & \checkmark & 36.8 & $3.22\pm1.57$ & $2.37\pm1.57$ & -- & -- \\
\hline
DMLVS~\cite{ICRA2023Felton}$^\dagger$ & $\times$ & 100.0 & $0.00\pm0.00$ & $0.00\pm0.00$ & -- & -- \\
DMLVS~\cite{ICRA2023Felton}$^\dagger$ & \checkmark & 76.0 & $1.93\pm1.28$ & $1.92\pm1.28$ & -- & -- \\
\hline\hline
\multicolumn{7}{l}{\textit{Training-free: pixel-level descriptors}} \\
\hline
SIFT-IBVS & $\times$ & 94.8 & $0.06\pm0.03$ & $0.05\pm0.02$ & -- & 2.9 \\
SIFT-IBVS & \checkmark & 38.8 & $0.29\pm0.62$ & $0.25\pm0.51$ & -- & 2.8 \\
\hline
AKAZE-IBVS & $\times$ & 90.6 & $0.03\pm0.01$ & $0.03\pm0.01$ & -- & 2.1 \\
AKAZE-IBVS & \checkmark & 64.8 & $0.09\pm0.06$ & $0.08\pm0.06$ & -- & 1.7 \\
\hline
ORB-IBVS & $\times$ & 95.4 & $0.17\pm0.11$ & $0.13\pm0.10$ & -- & 22.4 \\
ORB-IBVS & \checkmark & 48.0 & $0.46\pm3.68$ & $0.39\pm3.24$ & -- & 20.8 \\
\hline\hline
\multicolumn{7}{l}{\textit{Training-free: patch-level ViT}} \\
\hline
ViT-VS: DINOv2-308~\cite{ViTVS2025} & $\times$ & 100.0 & $1.86\pm1.07$ & $1.50\pm0.78$ & 1254 & 15.1 \\
ViT-VS: DINOv2-308~\cite{ViTVS2025} & \checkmark & 76.6 & $2.15\pm1.21$ & $1.83\pm0.98$ & 1230 & \underline{14.4} \\
\hline
ViT-VS: DINOv2-518 & $\times$ & 100.0 & $1.04\pm0.50$ & $0.81\pm0.38$ & 1328 & 6.3 \\
ViT-VS: DINOv2-518 & \checkmark & 77.8 & $2.60\pm1.17$ & $2.22\pm1.00$ & 1270 & 6.1 \\
\hline
ViT-VS: DINOv3-1440 & $\times$ & 100.0 & $\underline{0.52\pm0.19}$ & $\underline{0.44\pm0.17}$ & 2430 & 1.1 \\
ViT-VS: DINOv3-1440 & \checkmark & 81.0 & $1.65\pm1.00$ & $1.43\pm0.88$ & 2431 & 0.9 \\
\hline
\rowcolor[gray]{0.92}
\textbf{ART-VS: DINOv2-7$\times$7} & $\times$ & 100.0 & $0.74\pm0.39$ & $0.64\pm0.31$ & 2172 & 15.9 \\
\rowcolor[gray]{0.92}
\textbf{ART-VS: DINOv2-7$\times$7} & \checkmark & 87.2 & $0.88\pm0.67$ & $0.78\pm0.63$ & 2174 & 10.5 \\
\hline
\rowcolor[gray]{0.92}
\textbf{ART-VS: DINOv3-6$\times$6} & $\times$ & 100.0 & $0.75\pm0.36$ & $0.61\pm0.32$ & 1766 & \textbf{16.4} \\
\rowcolor[gray]{0.92}
\textbf{ART-VS: DINOv3-6$\times$6} & \checkmark & $\mathbf{95.4}$ & $1.01\pm0.61$ & $0.88\pm0.56$ & 1767 & 11.9 \\
\hline
\rowcolor[gray]{0.92}
\textbf{ART-VS: AM-Radio-6$\times$6} & $\times$ & 100.0 & $0.87\pm0.46$ & $0.74\pm0.38$ & 2778 & 15.8 \\
\rowcolor[gray]{0.92}
\textbf{ART-VS: AM-Radio-6$\times$6} & \checkmark & 91.0 & $\mathbf{0.81\pm0.47}$ & $\mathbf{0.69\pm0.44}$ & 2779 & 6.8 \\
\hline
\end{tabular}
\vspace{-2ex}
\end{table*}

We evaluate ART-VS in simulation (Sec.~\ref{sec:results}) with ablation studies on the phase transition threshold and ROI components (Sec.~\ref{sec:ablations}), followed by end-effector positioning and category-level grasping experiments (Sec.~\ref{sec:realworld}).

\subsection{Experimental Setup}
\label{sec:setup}
\textbf{Simulation.}
We replicate the simulation setup of~\cite{ICRA2023Felton, ViTVS2025}: a virtual Intel RealSense D435i captures a 60$\times$80\,cm poster at 1440$\times$1080 resolution in Gazebo, with robot control managed through ROS.
We sample 500 initial camera poses from a 1.2$\times$1.2$\times$0.3\,m volume centered 0.6\,m from the poster, with look-at points on four concentric circles (8 - 32\,cm radii) and random focal-axis rotation within $[-120°, 120°]$.
Mean initial errors are 46.4$\pm$17.0\,cm (position) and 74.1$\pm$27.7° (orientation).
Trials terminate upon convergence or after 1500 iterations.
Perturbed trials apply color-jitter (brightness 0.6, contrast 0.4), random erasing (probability 0.5, scale 0.02--0.33), and Gaussian blur ($\sigma$=0.05) to the target object, simulating appearance variations.
Computation runs on an NVIDIA RTX 4060 Ti GPU with Intel Core i7-13700K CPU.

\textbf{Real-World Hardware.}
A Universal Robots UR5 with wrist-mounted Intel RealSense D435 (1440$\times$1080) and Robotiq 2F-85 gripper is controlled via ROS in eye-in-hand configuration.
Computation runs on an NVIDIA RTX 4070 mobile GPU with AMD Ryzen 9 7940HS CPU.

\textbf{Convergence Criterion.}
Following~\cite{ICRA2023Felton}, a trial is considered converged if the pose error is reduced by $>$90\% relative to the initial displacement and commanded end-effector velocities fall below a near-zero threshold. 
Position and orientation errors are reported only for converged trials.

\textbf{Baselines.}
We compare against training-free classical descriptors (SIFT~\cite{SIFT}, ORB~\cite{ORB}, AKAZE~\cite{AKAZE}) and ViT-VS method~\cite{ViTVS2025} (DINOv2-Small at $308{\times}308$).
To isolate the effect of ViT processing resolution, we additionally evaluate the same single-pass ViT-VS scheme at two higher processing sizes: DINOv2-518 (DINOv2-Small, $518{\times}518$) and DINOv3-1440 (DINOv3-Small, $1440{\times}1080$).
Classical methods use standard OpenCV matchers (FLANN with Lowe's ratio test for SIFT; brute-force with cross-check for ORB and AKAZE), while all ViT methods use best-buddy matching~\eqref{eq:best_buddy}.
All methods share the same IBVS control law~\eqref{eq:control_law}, gain, convergence criteria, and number of correspondences ($K{=}24$) on our $1440{\times}1080$ input.
We additionally report published results for scene-specific methods (PBVS-CNN~\cite{ICRA2018Bateux}, DMLVS~\cite{ICRA2023Felton}).

\textbf{Implementation Details.} For DINOv2-Small (ViT-S/14) and DINOv3-Small (ViT-S/16), features are extracted from the final transformer block; for AM-RADIO (C-RADIOv3-B, ViT-B/16), the native spatial feature output is used. 
Following~\cite{amir2022deep}, hierarchical feature binning ($\beta$=1)  aggregates each patch descriptor with its 3$\times$3 neighborhood into 9-bin  descriptors.
From the set of mutually consistent correspondences~\eqref{eq:best_buddy}, $K$=24 are randomly subsampled per iteration.
Control gain is $\lambda$=0.2 for simulation, and 0.1 for robotic experiments. Velocity commands are smoothed with an exponential moving average ($\alpha$=0.8)~\cite{ViTVS2025}.

\subsection{Simulation Results}
\label{sec:results}

Table~\ref{tab:hollywood} presents results on the Hollywood poster benchmark across 500 trials.
We focus our analysis on comparisons among training-free methods; scene-specific methods (marked $^\dagger$) are included for reference.

\textbf{Convergence and Robustness.}
Under perturbation, ART-VS with DINOv3 achieves 95.4\% convergence - an 18.8 percentage point improvement over ViT-VS~\cite{ViTVS2025} (76.6\%) and 14.4 points over full-resolution DINOv3-1440 (81.0\%).
Among scene-specific methods, DMLVS~\cite{ICRA2023Felton} achieves perfect convergence when unperturbed but drops to 76.0\% under perturbation - comparable to ViT-VS~\cite{ViTVS2025} - while PBVS-CNN~\cite{ICRA2018Bateux} degrades to 36.8\%.
Classical descriptors degrade severely under perturbation (SIFT: 94.8\%$\to$38.8\%, AKAZE: 90.6\%$\to$64.8\%), while all three ART-VS variants remain above 87\%, confirming that the adaptive tiling strategy generalizes across ViT backbones.
Notably, progressively increasing single-pass resolution from DINOv2-308 through DINOv2-518 to DINOv3-1440 yields only incremental robustness gains (76.6\%→77.8\%→81.0\%) despite up to 12$\times$ more feature patches, whereas ART-VS's phase-aware resolution switching yields a substantially larger improvement.
This is consistent with the phase-dependent resolution trade-off motivating ART-VS: fine-grained features improve positioning only once the viewpoint gap is small, while at large pose differences, coarse patch aggregation provides more stable correspondences.

\textbf{Positioning Precision.}
ART-VS (DINOv3-6${\times}$6) reduces mean position error under perturbation by 53\% relative to ViT-VS~\cite{ViTVS2025} ($1.01{\pm}0.61$\,cm vs. $2.15{\pm}1.21$\,cm) with 50\% lower variance. Against the matched backbone at full resolution (DINOv3-1440: $1.65{\pm}1.00$\,cm), the reduction is 39\%.
In unperturbed conditions, full-resolution DINOv3-1440 achieves slightly better precision (0.52 vs.\ 0.75\,cm) - a trade-off we consider acceptable given the rarity of perturbation-free conditions in deployment.

\textbf{Computational Cost.}
We report effective FPS as total iterations divided by wall-clock time to convergence, naturally weighting the coarse and tiled phases by their respective duration.
ART-VS with DINOv3 achieves 16.4 effective FPS unperturbed and 11.9 under perturbation, compared to 1.1 and 0.9~FPS for DINOv3-1440 - over 13$\times$ faster at 27\% lower VRAM (1767 vs.\ 2431\,MB).
The lower effective rate under perturbation reflects longer convergence trajectories that spend a larger fraction of iterations in the tiled phase.
The higher control frequency enables faster correction of transient matching errors, contributing to the convergence advantage over full-resolution processing.

\begin{figure}[h]
    \vspace{1ex}
    \centering
    \includegraphics[width=1\linewidth]{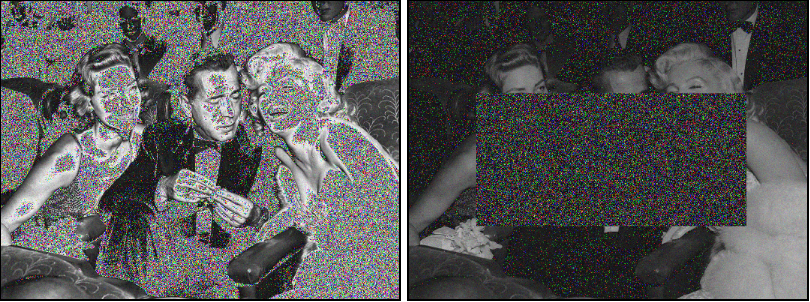}
    \caption{\textbf{Perturbed target images:} from two representative non-converged ART-VS DINOv3-6×6 trials (23/500, 4.6\%), showing dense noise with color jitter (left) and combined occlusion with noise (right).}
    \label{fig:failure_analysis}
\end{figure}

\textbf{Failure analysis.}
We analyze the 23 non-converged trials (4.6\%) of ART-VS with DINOv3-6×6, our best-performing configuration under perturbation (Fig. \ref{fig:failure_analysis}).
The dominant failure mode (17/23) is severe feature corruption from simultaneous large-area occlusion and dense noise, preventing meaningful error reduction.
Four trials achieved $>$90\% error reduction before perturbation-corrupted correspondences caused slow drift.
The remaining two trials diverged within the first iterations due to incorrect discrete rotation alignment under heavy noise. 
Since all 500 trials are drawn from the same initial pose distribution, failures are attributed to perturbation severity rather than initial pose difficulty.

\subsection{Ablation Studies}
\label{sec:ablations}

\begin{figure}[h]
    \centering
    \includegraphics[width=0.9\linewidth]{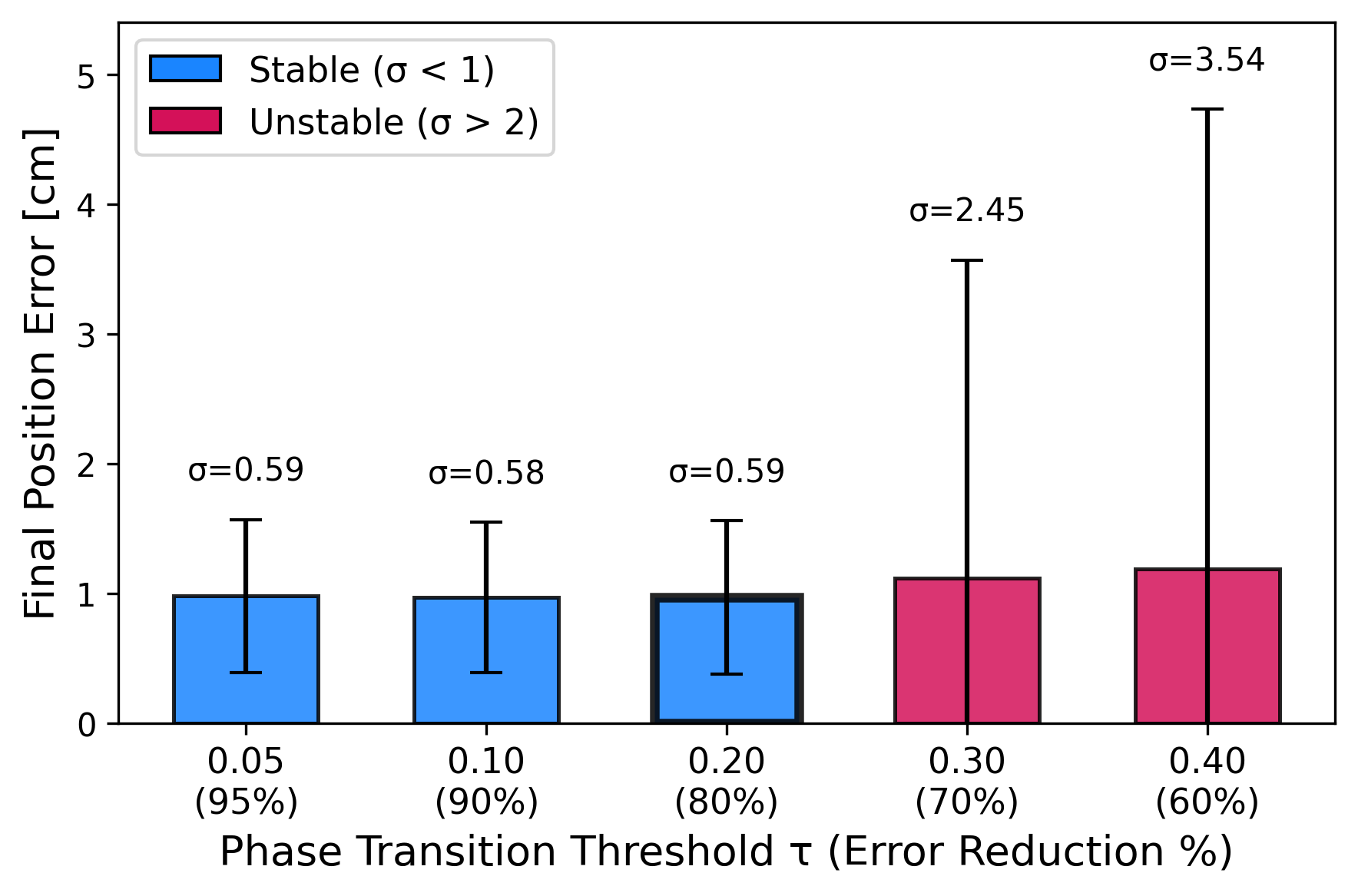}
    \caption{\textbf{Phase transition threshold ablation}. Early switching ($\tau \geq 0.30$) increases error and variance; late switching ($\tau \leq 0.10$) offers no precision gain. $\tau{=}0.20$ yields the lowest median error and lowest variance.}
    \label{fig:threshold_ablation}
\end{figure}

\textbf{Phase Transition.}
Fig.~\ref{fig:threshold_ablation} evaluates the coarse-to-fine transition threshold $\tau$ under perturbation on the simulation benchmark (500 trials).
Early switching ($\tau \geq 0.30$) activates the tiled refinement phase while pose differences are still large enough that corresponding content falls in different tiles across views, increasing position variance by up to 4$\times$ (0.59\,cm $\to$ 2.45\,cm at $\tau{=}0.30$).
Late switching ($\tau \leq 0.10$) matches the precision of $\tau{=}0.20$ but prolongs the coarse phase unnecessarily.
The selected threshold $\tau{=}0.20$ achieves the best trade-off: the earliest transition point at which tile correspondences remain reliable, yielding the lowest error and variance.
While $\tau$ was selected in simulation, real-world tuning of such thresholds is generally impractical; the positioning and grasping experiments in Sec.~\ref{sec:realworld} confirm that this value transfers without adjustment.

\textbf{ROI Detection and Tracking.}
The optional ROI-guided mode is used exclusively in robotic grasping experiment (Sec.~\ref{sec:realworld}); the simulation benchmark operates on full images.
We therefore evaluate tracking components separately on YCB-Video~\cite{ycbvideo}, which provides ground-truth annotations for multi-object scenes.
Since ART-VS requires only a single detection at initialization followed by per-frame tracking, we assess propagation quality across three trackers (BBox CSRT~\cite{BBoxCSRT}, Optical Flow~\cite{OpticalFlow}, LightTrack~\cite{LightTrack}) paired with two promptable detectors (LangSAM~\cite{xu2025langsam}, SAM3~\cite{sam3}) over 320 tracking instances across 67 videos.

Table~\ref{tab:ycb_tracking} reports IoU and Dice scores.
Performance is primarily determined by the choice of tracker: LightTrack achieves 0.67$\pm$0.20 IoU with LangSAM versus 0.50 for BBox CSRT and 0.15 for Optical Flow.
Detector choice has negligible impact (LangSAM+LightTrack: 0.67 vs.\ SAM3+LightTrack: 0.68), so we select LangSAM for its native language-promptable interface, which enables target specification via natural language without predefined class labels or manual geometric prompts.
LightTrack adds negligible overhead at 102\,FPS; LangSAM's 0.53\,s inference is a one-time initialization cost.

\begin{table}[h!]
\caption{ROI detection and tracking ablation on YCB-Video~\cite{ycbvideo} (320 instances). IoU and Dice as $\mu \pm \sigma$.}
\label{tab:ycb_tracking}
\centering
\begin{tabular}{lcc}
\toprule
Detector + Tracker & IoU & Dice \\
\midrule
LangSAM + LightTrack   & \textbf{0.67 $\pm$ 0.20} & \textbf{0.77 $\pm$ 0.18} \\
LangSAM + BBox CSRT        & 0.50 $\pm$ 0.17 & 0.63 $\pm$ 0.18 \\
LangSAM + Optical Flow & 0.15 $\pm$ 0.10 & 0.23 $\pm$ 0.13 \\
SAM3 + LightTrack      & \textbf{0.68 $\pm$ 0.18} & \textbf{0.78 $\pm$ 0.16} \\
SAM3 + BBox CSRT           & 0.51 $\pm$ 0.17 & 0.64 $\pm$ 0.17 \\
SAM3 + Optical Flow    & 0.15 $\pm$ 0.10 & 0.22 $\pm$ 0.13 \\
\bottomrule
\end{tabular}
\end{table}

\subsection{Robotic Experiments}
\label{sec:realworld}

We demonstrate ART-VS on a UR5 manipulator in two settings: end-effector positioning to assess convergence behavior, and category-level grasping to evaluate generalization under clutter.
All robotic experiments use the DINOv3-Small backbone with $\lambda{=}0.1$.

\begin{figure}[h]
\centering
\vspace{1ex}
\includegraphics[width=0.90
\linewidth]{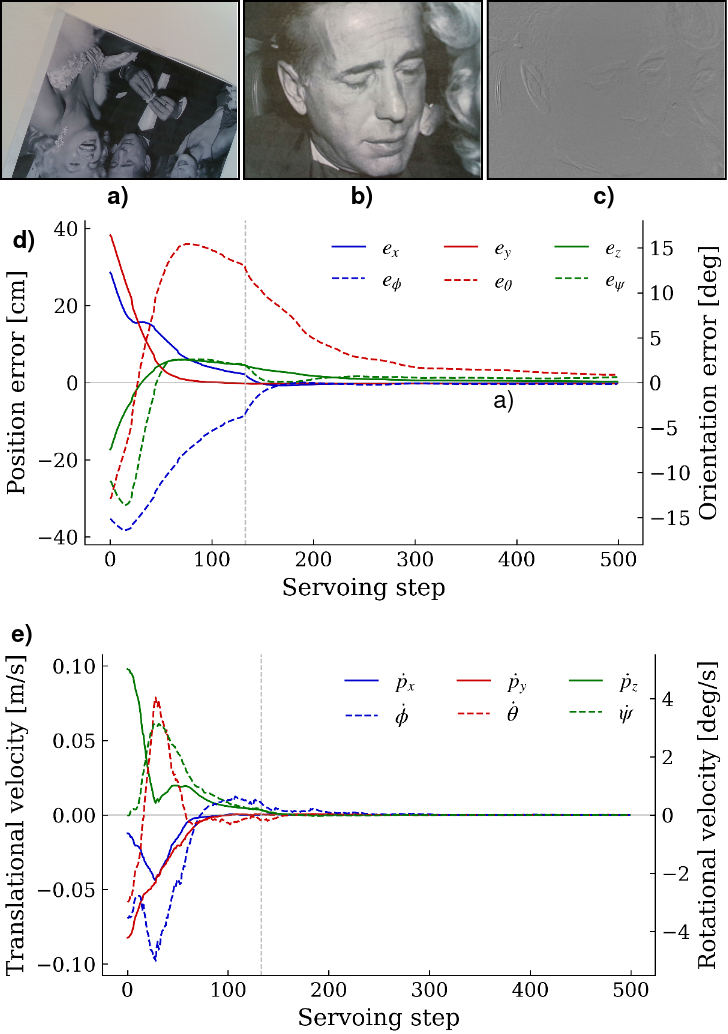}
\caption{\textbf{Real-world positioning trial.} a)~Initial image. b) ~Desired image I*. c)~Difference between desired I* and final image I. d)~Pose error over iterations. e)~Camera velocities.}
\label{fig:realrobot_detaile}
\vspace{-3ex}
\end{figure}

\textbf{End-Effector Positioning}
Fig.~\ref{fig:realrobot_detaile} shows a representative trial from a challenging initial displacement of 28.5\,cm depth, 38.2\,cm horizontal, and 17.3\,cm vertical offset with $-$12.6° roll, $-$13.1° pitch, and 160.1° yaw (yaw partially resolved by rotation compensation prior to servoing).
ART-VS transitions to the tiled refinement phase at iteration~133, after which the system refines to a final error of $-$0.06\,cm depth, $-$0.13\,cm horizontal, and 0.18\,cm vertical with $-$0.18° roll, 0.54° pitch, and 0.37° yaw - corresponding to 99.5\% position and 99.6\% orientation error reduction.
Positioning errors (Fig.~\ref{fig:realrobot_detaile}d) decrease monotonically during the coarse phase, with the phase transition clearly visible as a brief transient before continued convergence.

\begin{figure}[h]
    \centering
    \includegraphics[width=1\linewidth]{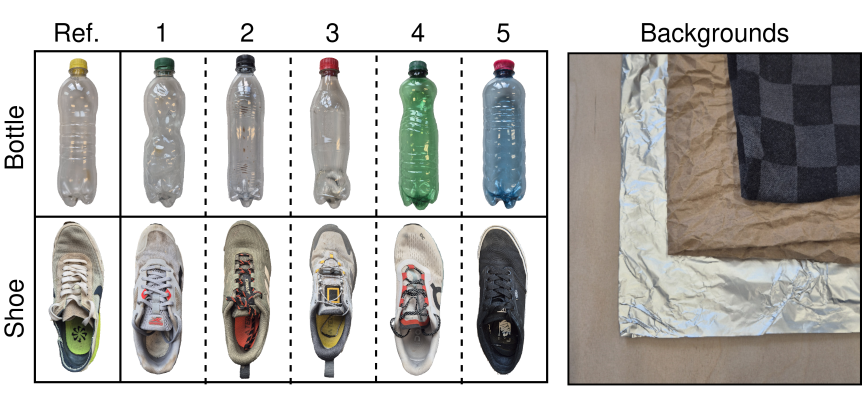}
    \caption{\textbf{Grasping objects and backgrounds.} Reference objects (left) and five test instances (1-5). Backgrounds (right): plain wood, aluminium foil, crumpled kraft paper, textured kitchen towel.}
    \label{fig:object_gallery}
\end{figure}

\begin{figure}[h]
    \centering
    \includegraphics[width=0.9\linewidth]{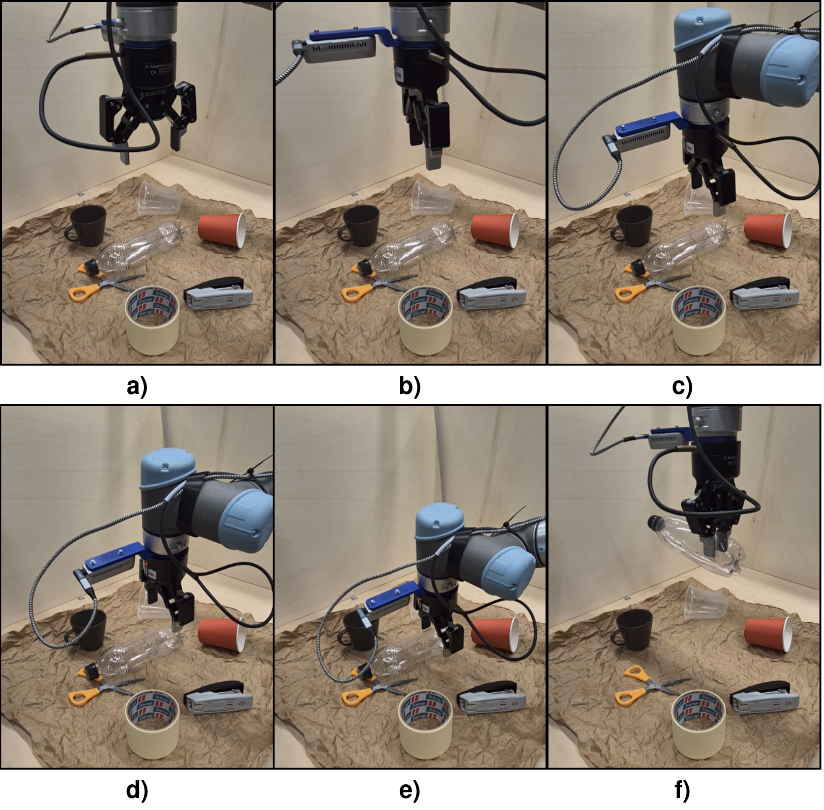}
    \caption{\textbf{Grasping sequence of a transparent bottle.} a)~Initial pose. b)~After rotation compensation. c)~After coarse phase. d)~After tiled refinement phase. e)~Grasping position. f)~Lifted object.}
    \label{fig:grasping_sequence}
    \vspace{-1ex}
\end{figure}

\textbf{Category-Level Grasping}
We evaluate category-level generalization on two challenging object categories - transparent PET bottles and shoes - using a single reference image per category to servo toward five visually distinct test instances (Fig.~\ref{fig:object_gallery}, left).
ART-VS operates in ROI-guided mode with text prompts ``transparent bottle'' or ``shoe.''
Each of the five instances per category is tested across four backgrounds of increasing visual complexity (plain wood, textured kitchen towel, crumpled kraft paper, aluminium foil; Fig.~\ref{fig:object_gallery} right) with six category-different distractors present in every trial.
We conduct 100 trials per category (5 instances $\times$ 4 backgrounds $\times$ 5 poses), with initial poses sampled from a $0.2{\times}0.2{\times}0.1$\,m workspace at 0.45\,m elevation with $\pm$120° roll variation (mean initial error: 8.86$\pm$2.45\,cm, 70.19$\pm$23.81°).
Upon convergence, the end-effector descends along the camera $Z$-axis and executes a parallel-jaw grasp; a trial is successful if the object is lifted. Fig.~\ref{fig:grasping_sequence} shows a representative bottle grasping sequence.

\begin{table}[h!]
\vspace{-1ex}
\caption{Category-level grasping results.}
\label{tab:realworld}
\centering
\setlength{\tabcolsep}{3.5pt}
\begin{tabular}{l|ccccc|c}
\toprule
 & \multicolumn{5}{c|}{Instance} & \\
\cmidrule(lr){2-6}
Category & 1 & 2 & 3 & 4 & 5 & Total \\
\midrule
Bottles  & 19/20 & 19/20 & 20/20 & 19/20 & 18/20 & \textbf{95/100} \\
Shoes & 20/20 & 20/20 & 19/20 & 20/20 & 19/20 & \textbf{98/100} \\
\bottomrule
\end{tabular}
\vspace{-1ex}
\end{table}

Table~\ref{tab:realworld} presents per-instance results.
ART-VS achieves 95/100 on transparent bottles and 98/100 on shoes.
The five bottle failures stem from two causes: incorrect rotation compensation due to near-symmetric bottle appearance (3 cases) and late-stage divergence from specular reflections altering correspondences near convergence (2 cases).
The two shoe failures are caused by ambiguous matching from an extreme initial viewpoint (1 case) and progressive tracker drift shrinking the bounding box below a sufficient feature count (1 case).
Failures concentrate on the most challenging background (aluminium foil: 3 of 7 total), consistent with the perturbation sensitivity observed in simulation.

\section{CONCLUSION}

We present ART-VS, an adaptive resolution method for training-free ViT-based VS that exploits a key observation: the utility of feature resolution is phase-dependent.
Coarse patch descriptors provide the robustness needed for initial convergence under appearance variation, while fine-grained features enable precise positioning only once the pose difference is small enough for spatially constrained matching.
The convergence rate improvement when simply increasing the input resolution for matching is marginal (+4.4\,pp), whereas our adaptive tiling yields a substantially larger gain (+18.8\,pp) over the same baseline, while maintaining over 10$\times$ higher effective FPS than full-resolution processing.
Hence, convergence robustness benefits from spatially constraining matching windows. 
Real-world grasping experiments across 200 trials demonstrate reliable category-level manipulation of unseen object instances from a single reference image, validating that the approach generalizes beyond controlled benchmarks.

Our failure analysis reveals two primary limitations: near-symmetric objects degrade rotation compensation, and severe simultaneous perturbations corrupt features beyond recovery. Addressing viewpoint symmetry through symmetry-aware matching is a key direction for improving robustness. Future work will also investigate RGB-only operation through monocular depth estimation and integrate language grounding directly into semantic feature extraction, eliminating the need for separate detection and tracking components.

\bibliographystyle{IEEEtran}
\bibliography{refs}
\end{document}